\pdfoutput=1

\documentclass[11pt]{article}
 
\usepackage[preprint]{acl}
\usepackage{fix-cm}

\usepackage{times}
\usepackage{latexsym}

\usepackage{amsmath,amsfonts,bm}

\def\eqref#1{equation~\ref{#1}}

\def\1{\bm{1}}

\def\va{{\bm{a}}}

\def\vx{{\bm{x}}}

\def\mK{{\bm{K}}}

\def\mQ{{\bm{Q}}}
\def\mR{{\bm{R}}}

\def\mW{{\bm{W}}}
\def\mX{{\bm{X}}}
\def\mY{{\bm{Y}}}

\DeclareMathAlphabet{\mathsfit}{\encodingdefault}{\sfdefault}{m}{sl}
\SetMathAlphabet{\mathsfit}{bold}{\encodingdefault}{\sfdefault}{bx}{n}

\def\gL{{\mathcal{L}}}

\newcommand{\E}{\mathbb{E}}

\usepackage{multirow}
\usepackage{rotating}
\usepackage[T1]{fontenc}

\usepackage[utf8]{inputenc}

\usepackage{microtype}

\usepackage{inconsolata}

\usepackage{graphicx}

\usepackage{tikz}
\usetikzlibrary{positioning}
\usetikzlibrary{calc}
\usetikzlibrary{shapes.geometric}
\usepackage{caption}    
\usepackage{caption}    

\usepackage{booktabs}
\usepackage{multirow}
\usepackage{amssymb}  
\usepackage{pifont}   
\usepackage{arydshln}
\usepackage{graphicx}
\usepackage{amsmath}
\usepackage{amssymb}
\usepackage{mathtools}
\usepackage{amsthm}
\usepackage{nicefrac}       

\usepackage[capitalize,noabbrev]{cleveref}

\theoremstyle{plain}
\newtheorem{theorem}{Theorem}[section]

\theoremstyle{definition}
\newtheorem{definition}[theorem]{Definition}

\theoremstyle{remark}

\newcommand{\method}{KurTail\ }

\title{\method: Kurtosis-based LLM Quantization}

\author{
 \textbf{Mohammad Sadegh Akhondzadeh\textsuperscript{1}\thanks{Work done during an internship at Axelera AI}},
 \textbf{Aleksandar Bojchevski\textsuperscript{1}},\\
 \textbf{Evangelos Eleftheriou\textsuperscript{2}},
 \textbf{Martino Dazzi\textsuperscript{2}}
\\
\\
 \textsuperscript{1}University of Cologne,
 \textsuperscript{2}Axelera AI,
\\
 \small{
   \textbf{Correspondence:} \href{mailto:akhondzadeh@cs.uni-koeln.de}{akhondzadeh@cs.uni-koeln.de}
 }
}

\begin{document}
\maketitle
\begin{abstract}
One of the challenges of quantizing a large language model (LLM) is the presence of outliers. Outliers often make uniform quantization schemes less effective, particularly in extreme cases such as 4-bit quantization. We introduce KurTail, a new post-training quantization (PTQ) scheme that leverages Kurtosis-based rotation to mitigate outliers in the activations of LLMs. Our method optimizes Kurtosis as a measure of tailedness. This approach enables the quantization of weights, activations, and the KV cache in 4 bits. 
We utilize layer-wise optimization, ensuring memory efficiency.
\method outperforms existing quantization methods, offering a 13.3\% boost in MMLU accuracy and a 15.5\% drop in Wiki perplexity compared to QuaRot \citep{QuaRot}. It also outperforms SpinQuant \citep{SpinQuant} with a 2.6\% MMLU gain and reduces perplexity by 2.9\%, all while reducing the training cost.
For comparison, learning the rotation using SpinQuant for Llama3-70B requires at least four NVIDIA H100 80GB GPUs, whereas our method requires only a single GPU, making it a more accessible solution for consumer GPU. 
\end{abstract}

\section{Introduction}
Large language models (LLMs) have advanced significantly in recent years, showcasing remarkable performance and capabilities. As these models grow in size and complexity, the computational cost required for their deployment and inference has increased dramatically. Furthermore, with new methods for inference time compute \citep{openai2024learning, guo2025deepseek}, enhancing inference speed (tokens per second) has become increasingly important. 
This has shifted the focus toward accelerating model performance while reducing memory and computational requirements. An effective method to achieve this is post-training quantization (PTQ), which involves representing model weights and/or activations in lower numerical precisions. PTQ can significantly reduce the memory footprint and computational overhead and subsequently decrease latency and energy consumption, which are especially beneficial for inference on resource-constrained edge devices.

Serving a model involves two stages of \textit{prefilling} and \textit{generation}. During \textit{prefilling}, the model processes the input prompt and stores the internal state, which is known as key-value (KV) caching. During \textit{generation}, tokens are produced auto-regressively. The \textit{prefilling} stage is considered compute-bound, while the generation stage is memory-bound due to repeated access to and updates of the KV cache. Quantizing each stage offers distinct advantages for improving inference efficiency. KV-cache quantization reduces memory requirements and accelerates data movement, which enhances the \textit{generation} stage, particularly in scenarios involving long-context inference. Weight quantization, on the other hand, reduces the memory footprint independently, and when it is combined with activation quantization, it also reduces the computational demands, which mainly speeds up the \textit{prefilling} stage. 
However, activation quantization presents challenges due to large outliers in certain channels \citep{llmint8, smoothquant}, which limits the effectiveness of uniform integer quantization as it destroys the dynamic range of the activations.
While channel-wise quantization can effectively address this issue, the lack of hardware support makes it computationally expensive in practice. 

 Several methods have been proposed to address this challenge. \citet{llmint8} and \citet{quik} advocate for mixed-precision computation in which they store some of the channels in higher precision and less sensitive channels in lower precision to balance accuracy and efficiency. \citet{smoothquant} introduces channel-wise scaling into the layer normalization and the weights of linear layers. \citet{QuaRot} proposed random rotation which takes the advantage of the computational invariance framework \citep{slicegpt} to mitigate the outliers problem.

We introduce \emph{\method} -- a novel approach to mitigating activation outliers by applying a learnable rotation to the activations. We advocate for learnable rotation instead of random rotation, which is suboptimal \citep{SpinQuant}. Unlike SpinQuant \citep{SpinQuant}, which requires end-to-end training of the model’s loss,
\method focuses on reducing the tail density of activations independently per layer. We perform layer-wise inference to store activations and optimize the transformation based on the Kurtosis of activations. As a result, \method can be implemented in a significantly more memory-efficient manner. For instance, while SpinQuant requires at least four NVIDIA H100 80GB GPUs to compute rotations for Llama3-70B, \method achieves the same with just a single GPU. Despite its lower computational requirements, \method outperforms existing methods in terms of perplexity and zero-shot reasoning tasks. \method outperforms existing quantization methods with a 13.3\% increase in MMLU accuracy and a 15.5\% decrease in Wiki perplexity compared to QuaRot\citep{QuaRot}. It also performs better than SpinQuant\citep{SpinQuant}, achieving a 2.6\% increase in MMLU accuracy and a 2.9\% decrease in perplexity, all while reducing the cost of training the rotation.

\section{Background}

\paragraph{Post Training Quantization.}
Previous work on post-training quantization fits into two main groups: weight-only quantization  \citep{gptq, awq, egiazarian2024extreme, quip} and weight-activation quantization \citep{smoothquant, llmint8, QuaRot, SpinQuant}. In weight only quantization, the weight are projected into a lower precision, such as 4 bits, 3 bits, or even less, and then de-quantized to higher
precision before the actual computation, with all calculations
still being done in high precision. Several studies \citep{smoothquant, QuaRot, SpinQuant} attempted to
introduce quantization methods for both weight and activation. They showed that uniform quantizing is impractical for large language models since they suffer from large outliers. To address
this issue, \citet{llmint8} proposed a mixed-precision approach for
handling outliers at higher precision. Others \citep{smoothquant, awq} proposed trading outliers between weights and activations by introducing a re-scaling paradigm.
\citet{quip} introduced an incoherence processing method using random rotation matrices and applying vector quantization on the weights for compression, adding overhead to inference. QuaRot \citep{QuaRot} was inspired by \citet{quip} and took advantage of the invariance framework proposed by \citet{slicegpt} introducing a rotation-based approach to compress and remove outliers from
the activation space using a random Hadamard rotation.
Later, SpinQuant \citep{SpinQuant} improves the results of QuaRot \citep{QuaRot} by optimizing some of these rotations to minimize cross-entropy loss through end-to-end training.
While SpinQuant improves the results compared to QuaRot it suffers from a high computational cost for learning the rotations. In this work, we attempt to address this issue by introducing a novel approach for learning the rotations.

\paragraph{Uniform Quantization for $k$-bit Precision.}
\label{sec: uniform_qunatization}
For a given vector  $\vx$, uniform integer quantization reduces its continuous range of values to a finite set of discrete levels, enabling representation in lower precision. In $ k$-bit quantization, the value range $[\vx_{\text{min}}, \vx_{\text{max}}]$ is divided into $ 2^k$ equal intervals. Each element $ \vx_{i}$  in $\vx$ is mapped to its closest quantization level by $ Q(x_{i}) = \text{round}\left(\frac{\vx_{i} - {b}}{s}\right) \cdot s + {b}$. Here $s$ is the scale factor or step size and $b$ is the shift. The values of $s$ and $b$ depend on the specific quantization scheme. In symmetric quantization, the range is assumed to be symmetric around zero. Therefore, $b = 0$, and 
$s = \frac{\max(|\vx_{\text{max}}|, |\vx_{\text{min}}|)}{2^{k-1} - 1}$.
Alternatively, in asymmetric quantization, the range is not assumed to be centered at zero and therefore, 
$ b = \min(\vx), \quad s = \frac{\vx_{\text{max}} - \vx_{\text{min}}}{2^k - 1}.$ 
Given $\vx$ sampled from a distribution $f$, quantizer $Q$ minimize the error between the quantized and the original values. The expected mean-squared error (MSE), is defined as: 
\begin{equation}
    \text{MSE}(\vx, Q) = \mathbb{E} \left[ \left( \vx - Q(\vx) \right)^2 \right]
\end{equation}

\begin{definition}
\label{def: sensitivity}
\label{app: defsensitivity}
\textbf{Quantization Sensitivity} \citep{kurtosis_weight}
    For a given distribution $f$ and its corresponding vector  $\vx$, let  $\tilde{s}$ denote the optimal quantization step size where $\tilde{s}$ minimizes the quantization error, and let $Q_{\tilde{s}}(\vx)$ represent the optimal quantizer. Quantization sensitivity $\Gamma(\vx, \epsilon)$  is defined as the increase in the mean squared error (MSE) caused by a small perturbation \( \epsilon > 0 \) in the quantization step size \( s \) around \( \tilde{s} \), such that \( |s - \tilde{s}| = \epsilon \). Specifically, the sensitivity is given by:
\begin{equation}
\Gamma(\vx, \epsilon) = \left| \text{MSE}(\vx, s) - \text{MSE}(\vx, \tilde{s}) \right|
\end{equation}
\end{definition}
\begin{theorem}
    \label{theorem: uni}
    \citep{kurtosis_weight} Considering $\vx_U$ and $\vx_N $ be continuous random variables with uniform and normal distributions. Then, for any given $\varepsilon > 0$, the quantization sensitivity $\Gamma(\vx, \varepsilon)$ satisfies $\Gamma(\vx_U, \varepsilon) < \Gamma(\vx_N, \varepsilon)$.
\end{theorem}
This theorem indicates that, compared to the typical normal distribution, the uniform distribution is more robust to changes in the quantization step size $s$. Therefore, it becomes apparent that there is great benefit in adjusting the distribution of the activations and weight to get closer to uniform distribution. This implies that the uniform distribution is a perfect fit for uniform quantization. It can also be shown for the uniform distribution the optimal scaling, $\tilde{s}$ is equal to $\quad s = \frac{x_{\text{max}} - x_{\text{min}}}{2^k - 1}$. 

\citet{kurtosis_weight} also show that the optimal step size for a uniform distribution closely approximates the most robust quantization (less sensitive) step size. 

\paragraph{Kurtosis.}
 Kurtosis is a statistical measure that describes the degree of tailedness in the distribution of a dataset. It helps determine whether the data have heavy or light tails compared to a normal distribution. Mathematically, Kurtosis is defined as the standardized fourth moment of a population around its mean, and it is calculated using 
\begin{equation}
     \kappa = \frac{\E[(x - \mu)^4]}{(\E[(x - \mu)^2])^2} = \frac{\mu_4}{\sigma^4}
\end{equation}
where $\mu$ is the mean, $\mu_4$ is the fourth moment about the mean, and $\sigma$ is the standard deviation. The Kurtosis of a normal distribution is 3.
To center the Kurtosis value at zero for the normal distribution, the adjusted measure \(\textrm{Kurtosis} - 3\) is often used, which is referred to as excess Kurtosis. 

Positive Kurtosis is characterized by heavy tails and a sharp peak (indicating greater tail density than a normal distribution, e.g., the Laplacian distribution). Positive Kurtosis also means the shift of mass from the shoulders to both the tails and the center. 
On the contrary, negative Kurtosis is a sign of light tails and a flatter distribution (like uniform or beta distribution) caused by mass moving from the tails and center to the shoulders. \citet{4bitqunatization_cnn} demonstrate that deep neural network weights and activations typically follow Gaussian or Laplace distributions. Furthermore, \citet{llmint8} identifies the presence of extreme outliers in LLM parameters, which are critical for maintaining performance. 

Our key insight is that distributions with outliers exhibit high kurtosis, which measures the presence of extreme values. Therefore, by optimising the rotation to minimize the kurtosis we can bring the distribution closer to uniform.

Uniform distribution is the desired distribution of the activations and weights for uniform quantization (\autoref{sec: uniform_qunatization}), as we aim to move the distribution closer to a uniform distribution. Kurtosis serves a two purpose: first, to encourage the distribution to resemble a uniform distribution, and second, to reduce the outliers by penalizing them. Therefore, we define the loss function as:
\begin{equation*}
\gL_{\kappa} =  \frac{1}{L} \sum_{i=1}^{L}|\kappa({{\bigoplus}_{j=1}^N\va_i}) - \kappa_u|
\end{equation*}
where  $\bigoplus$  denotes the concatenation of the activation of all tokens at that layer and $\kappa_u$ is the Kurtosis of the uniform distribution.    

\paragraph{Quantization Sensitivity.}
Quantization sensitivity measures the difference in the quantization error when we slightly perturb the optimal scaling \citep{kurtosis_weight}. Theoretically the sensitivity decreases as the distribution become closer to uniform (see \autoref{theorem: uni}). We evaluate our method by measuring activation sensitivity both before and after applying rotations optimized with Kurtosis. We expect that after applying these rotations, the activation distribution will be closer to uniform, resulting in better quantization robustness. We empirically measure the sensitivity of the activation distribution before and after applying the rotation. We utilize the Llama3.1 8-B model and apply two rotation techniques: one using a random Hadamard transformation and another using a Kurtosis-optimized rotation. First, we compute the optimal scaling \citep{kurtosis_weight} for activation quantization and then calculate the quantization sensitivity based on \autoref{def: sensitivity}.

\begin{figure}[ht]
    \resizebox{0.49\textwidth}{!}{
     \input{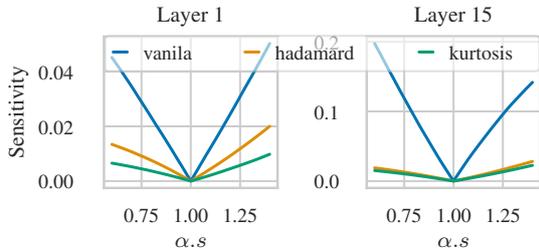}
     }
     \caption{Empirical sensitivity of the MHSA input distribution across different rotations.
     $\alpha$ indicates the fraction of the optimal step size used to analyze the sensitivity of quantization.
     }
     \label{fig: sensitivity}
\end{figure}

In \autoref{fig: sensitivity}, the symbol $\alpha$ indicates the fraction of the optimal step size used to analyze quantization sensitivity. The results show that the random Hadamard transformation reduces quantization sensitivity. Additionally, our Kurtosis-based method exhibits an even more significant reduction in sensitivity, suggesting that it more effectively aligns the distribution with uniformity. Interestingly, we also observed that the sensitivity drop is strongest in the first layer compared to other layers.  We also see a similar pattern for kurtosis. This difference is also noticeable in \autoref{fig: sensitivity} when we are comparing the first layer to layer 15. While we don’t show all layers, this trend holds for deeper layers.

\begin{figure}
    \resizebox{0.49\textwidth}{!}{
    \includegraphics{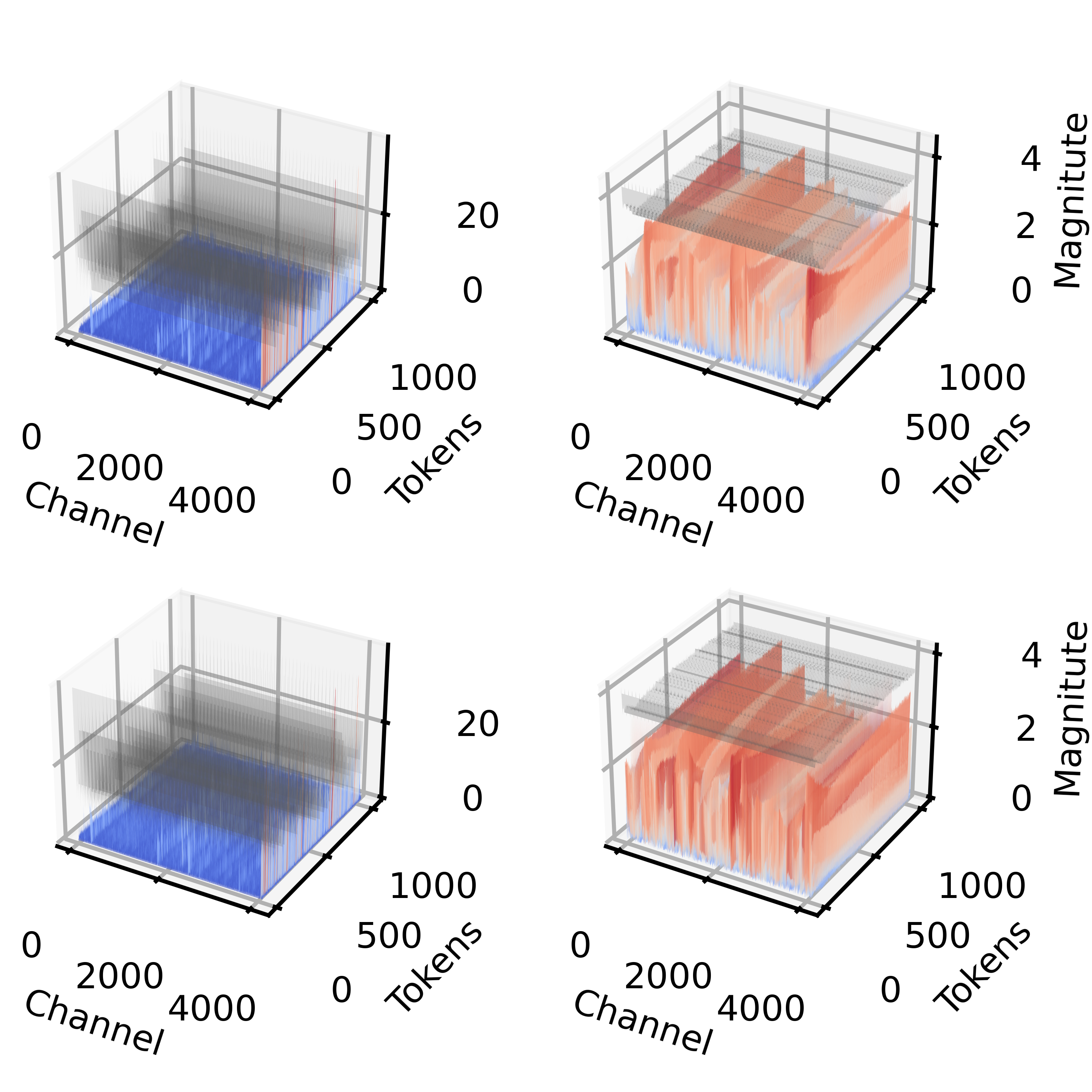}
    }
    \caption{The input distribution of the MHSA and FFN blocks in the LLaMA3-8B model is shown before and after applying \method.
    Before rotation, some channels have noticeable outliers, which can disrupt the data balance. The rotated distribution allows for more accurate token-wise qunatization.
    }
    \label{fig: 3d_outliers}
    
\end{figure}

\paragraph{Evaluation of \method on Channel Outliers.}

To demonstrate that the learned rotation by \method reduces the degree of tailedness in the distribution, we visualize the inputs of multi-head self-attention (MHSA) and feed-forward network (FFN) blocks of layer 15 in Llama3-8B. In \autoref{fig: 3d_outliers}, we compare the input distribution once without rotation and once with \method learned rotation. Additionally, we highlight the maximum value for each token with a gray surface above each token.
As shown, \method effectively mitigates outliers in activation quantization.

In dynamic per-token quantization, the maximum value of a token’s vector plays a critical role in determining the quantization step size and range. Larger maximum values increase the quantization range, which results in larger quantization steps and greater precision loss. Alternatively, reducing the maximum value allows for smaller quantization steps, which result in more efficient representation of token values with minimal degradation of information. Therefore, lowering the maximum values across tokens is directly connected to overall quantization error and model performance. To evaluate how well different methods achieve this goal, we measure the success rate of our proposed method, \method, compared to its un-rotated counterpart (baseline vector) and an alternative rotation method, QuaRot. A “success” is defined as a case where the maximum value of a token’s vector after applying a benchmark rotation method (\method or QuaRot) is smaller than that of the baseline vector. The success rate is defined as the percentage of tokens where the benchmarked rotated version achieves this reduction. In \autoref{tab:min_success}, we present the average success rates for LLAMA3-8B.  \method consistently produces smaller maximum values across all layers, samples, and tokens, achieving a higher success rate compared to the baseline vector in nearly all cases. Additionally, it outperforms QuaRot in approximately 63.29\% in MSHA, 62.99\% in FFN on average. 

\begin{table}[ht]
    \centering
    \caption{The success rate of benchmark over baseline.}
    \label{tab:min_success}
    \begin{tabular}{c|ll|c}
        \toprule
         & \textbf{\small{Baseline}} & \textbf{\small{Benchmark}} & \textbf{\small{Success Rate (\%)}} \\
        \midrule
        \multirow{3}{*}{\begin{turn}{90}MHSA\end{turn}} 
            & Vanilla & \method  & 99.74\% \\
            & Vanilla & QuaRot   & 99.43\% \\
            & QuaRot & \method   & 63.29\% \\
        \midrule
        \multirow{3}{*}{\begin{turn}{90}FFN \end{turn}} 
            & Vanilla & \method  & 99.96\% \\
            & Vanilla & QuaRot   & 99.96\% \\
            & QuaRot &  \method   & 62.99\% \\
        \bottomrule
    \end{tabular}
\end{table}

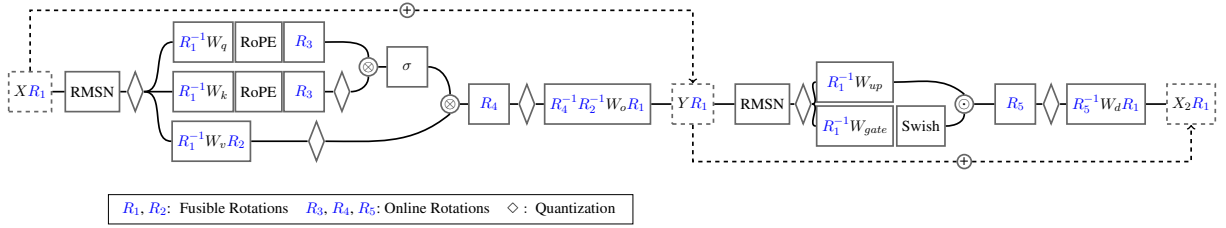
\begin{figure*}[t]
    \centering
    \resizebox{\textwidth}{!}{
\begin{tikzpicture}[
roundnode/.style={circle, draw=black!60, fill=white!5, very thick, minimum size=3mm, inner sep=0.2mm},
blacksquarednode/.style={rectangle, draw=black!60, fill=white!5, very thick, minimum size=10mm},
diamondnode/.style={diamond, draw=black!60, fill=white!5, very thick, minimum size=10mm, minimum width=0.4cm, inner sep=0.0},
]

\node[blacksquarednode, dashed] (XR){$X\textcolor{blue}{R_1}$};
\node[blacksquarednode] (rmsnorm1) [right=0.3cm of XR]{RMSN};

\draw[-, very thick] (XR.east) -- (rmsnorm1.west);

\node[diamondnode] (q1) [right=0.15cm of rmsnorm1]{};
\draw[-, very thick] (rmsnorm1.east) -- (q1.west);

\node[blacksquarednode] (k) [right=0.7cm of q1]{$\textcolor{blue}{R_1^{-1}}W_k$};
\node[blacksquarednode] (q) [above=0.2cm of k]{$\textcolor{blue}{R_1^{-1}}W_q$};
\node[blacksquarednode] (v) [below=0.2 of k, xshift=0.2cm]{$\textcolor{blue}{R_1^{-1}}W_v \textcolor{blue}{R_2}$};

\draw[-, very thick] (q1.east) -- (k.west);
\draw[-, very thick] (q1.east) to[out=0, in=180] (q.west);
\draw[-, very thick] (q1.east) to[out=0, in=180] (v.west);

\node[blacksquarednode] (Rope1) [right=0.01cm of k]{RoPE};
\node[blacksquarednode] (Rope2) [above=0.2cm of Rope1]{RoPE};

\node[blacksquarednode] (R3) [right=0.01cm of Rope1]{$\textcolor{blue}{R_3}$};
\node[diamondnode] (q_kv) [right=0.2cm of R3]{};
\node[blacksquarednode] (R3_2) [above=0.2cm of R3]{$\textcolor{blue}{R_3}$};
\node[diamondnode] (q2) [right=1.4cm of v]{};

\node[roundnode] (prod) at  ($(R3)!0.5!(R3_2) + (1.6cm,0)$){$\otimes$};

\draw[-, very thick] (R3.east) -- (q_kv.west);
\draw[-, very thick] (q_kv.east) to[out=0, in=-90] (prod.south);
\draw[-, very thick] (R3_2.east) to[out=0, in=90] (prod.north);

\node[blacksquarednode] (softmax) [right=0.2cm of prod]{$\sigma$};
\draw[-, very thick] (prod.east) -- (softmax.west);

\node[roundnode] (res1) [above=0.7 of softmax]{+};
\draw[-, very thick, dashed] (XR) |- (res1);

\node[roundnode] (prod2) at  ($(softmax)!0.5!(v) + (3.5cm,0)$){$\otimes$};
\draw[-, very thick] (softmax.east) to[out=0, in=90] (prod2.north);

\draw[-, very thick] (v.east) -- (q2.west);
\draw[-, very thick] (q2.east) to[out=0, in=-140] (prod2.south);

\node[blacksquarednode] (R4) [right=0.2cm of prod2]{$\textcolor{blue}{R_4}$};
\node[diamondnode] (q3) [right=0.2cm of R4]{};
\node[blacksquarednode] (out) [right=0.2 of q3]{$\textcolor{blue}{R_4^{-1}R_2^{-1}}W_o \textcolor{blue}{R_1}$};
\draw[-, very thick] (prod2) -- (R4);
\draw[-, very thick] (R4) -- (q3);
\draw[-, very thick] (q3) -- (out);

\node[blacksquarednode, dashed] (YR) [right=0.5 of out]{$Y\textcolor{blue}{R_1}$};
\draw[-, very thick] (out) -- (YR);

\draw[->,very thick, dashed] (res1) -| (YR);

\node[blacksquarednode] (rmsnorm2) [right=0.5cm of YR]{RMSN};
\node[diamondnode] (q_rms2) [right=0.1cm of rmsnorm2]{};

\draw[-, very thick](YR) -- (rmsnorm2);

\node[blacksquarednode, minimum width=1.9cm] (w_up) at ($(rmsnorm2)!0.5!(rmsnorm2) + (2.3cm,0.55)$){$\textcolor{blue}{R_1^{-1}}W_{up}$};
\node[blacksquarednode, minimum width=1.9cm] (w_down)  at ($(rmsnorm2)!0.5!(rmsnorm2) + (2.3cm,-0.55)$){$\textcolor{blue}{R_1^{-1}}W_{gate}$};

\draw[-, very thick] (rmsnorm2.east) -- (q_rms2.west);

\draw[-, very thick] (q_rms2.east) to[out=0, in=180] (w_up.west);
\draw[-, very thick] (q_rms2.east) to[out=0, in=180] (w_down.west);

\node[blacksquarednode] (swish)  [right=0.05cm of w_down]{Swish};

\node[roundnode] (dotprod) at  ($(w_up)!0.5!(swish) + (1.9cm,0)$){$\odot$};
\draw[-, very thick] (w_up.east) to[out=0, in=140] (dotprod.north);
\draw[-, very thick] (swish.east) to[out=0, in=-90] (dotprod.south);

\node[blacksquarednode] (R5)  [right=0.5 of dotprod]{$\textcolor{blue}{R_5}$};
\draw[-, very thick] (dotprod) -- (R5);

\node[diamondnode] (q4) [right=0.15cm of R5]{};
\draw[-, very thick] (R5) -- (q4);
\node[blacksquarednode] (Wd)  [right=0.15 of q4]{$\textcolor{blue}{R_5^{-1}}W_{d} \textcolor{blue}{R_1}$};
\draw[-, very thick] (q4) -- (Wd);

\node[blacksquarednode, dashed] (X2R)  [right=0.5 of Wd]{$X_2\textcolor{blue}{R_1}$};
\draw[-, very thick] (Wd) -- (X2R);

\node[roundnode](res2) [below=1 of dotprod]{+};
\draw[-,very thick, dashed] (YR) |- (res2);
\draw[->,very thick, dashed] (res2) -| (X2R);

\node at ($(current bounding box.south) + (-6,-1)$) [draw, fill=white, align=left] {
  \begin{tabular}{lll}
    \textcolor{blue}{$R_1$}, \textcolor{blue}{$R_2$}: \ Fusible Rotations& \textcolor{blue}{$R_3$},
    \textcolor{blue}{$R_4$},
    \textcolor{blue}{$R_5$}:\   Online Rotations & $\Diamond: $  \ Quantization
  \end{tabular}
};
\end{tikzpicture}
}
    \caption{Diagram of a single-layer decoder network after applying rotations. 
    Each block represents a computation unit.
    Blocks containing both blue and black indicate that the rotation is fused into the network without adding extra computation. In contrast, blocks with only the rotation signify additional computations during inference.}
    \label{fig: placement}
\end{figure*}
\section{\method}
\label{sec: method}
\paragraph{Placement of the Rotations.}
Following the computational invariance theorem — as introduced by \citet{elhage2023privileged, slicegpt} and later utilized by QuaRot and SpinQuant — we adopted a similar framework to transform the activation functions at each layer. The placement of rotations is illustrated in \autoref{fig: placement}. This figure depicts a single layer of a transformer model, where each square represents a computation block. The rotations are categorized into fusible rotations ($\mR_1$ and $\mR_2$) and online rotations ($\mR_3$, $\mR_4$, and $\mR_5$). Fusible rotations do not add additional computational costs during inference since they can be merged with the model’s original parameters. Specifically, we apply $\mR_1$ to the left side of the token embedding, $\mW_o$, and $\mW_d$ within the MHSA and FFN blocks, respectively. The inverse of $\mR_1$ is applied to the right side of $\mW_q$, $\mW_k$, $\mW_v$ in the attention block, and $\mW_{up}$, $\mW_{gate}$ in the FFN block. Due to the residual connection, the exact same rotation must also be applied across subsequent layers (e.g., $\mX\mR_1 + \mY\mR_1$ in one layer and $\mY\mR_1 + \mX_2\mR_1$ in the next). The second fusible rotation, $\mR_2$, is applied to the right side of $\mW_v$, with its inverse applied to the left side of $\mW_o$. This transformation improves the distribution of KV-caches and can vary across layers. The second group of rotations, $\mR_3$, $\mR_4$, and $\mR_5$, are online which minimally increase the computational costs compared to the original model but they improve the performance. To mitigate this, we utilize random Hadamard matrices, which are computationally efficient, resulting in minimal overhead. For $\mR_3$, the transformation is applied after each rotational positional encoding for queries and keys. Since the transpose of any orthogonal matrix equals its inverse, there is no need to add the inverse matrix explicitly. During the computation of attention scores, the term $\mQ^T\mK$ simplifies to $\mQ^T\mR_3^T\mR_3\mK$, effectively nullifying the impact of the rotation. For $\mR_4$, we introduce the transformation after applying the softmax scores to the values and add the inverse in the subsequent linear layer. Similarly, $\mR_5$ is implemented in the FFN block using the same approach.

\paragraph{Learning the Rotations.} To discover the optimal rotations, we first run the vanilla model and store the inputs from both the MHSA and FFN blocks. Next, we create a small network consisting of a linear layer and an RMSNorm, designed to simulate the inputs of the MHSA and FFN blocks before quantization (\autoref{fig: placement}). 

For optimization, we shuffle the stored input data from all transformer layers and both blocks and then train the rotation using Kurtosis loss. Since the optimization requires the rotations to remain within the orthogonal space, we use the Caley Adam \citep{li2020efficient} optimizer to enforce this constraint. We train this small network for 100 iterations using 500 samples from the WikiText \citep{merity2016wikitext} training set. In section \autoref{sec: cal_abl}, we also did an ablation study on the different calibration size and datasets. After training, the resulting rotation is fused into the original network. For the $\mR_2$, we apply a similar approach, but we removed the RMSNorm and just optimize the linear layer with the Kurtosis loss.\looseness=-1

\paragraph{Optimization in the Orthogonal Space.}
As discussed in \autoref{sec: method}, the transformation needs to be optimized in the orthogonal space to be consistent with a computational invariance framework. Therefore, we optimize all of the transformation matrices within the Stiefel Manifold \citep{li2020efficient} i.e., the space of orthonormal matrices, using Caley Stochastic Gradient Descent (SGD) or Caley Adam \citep{li2020efficient}. For more details see \citep{li2020efficient}.

\paragraph{Training Cost.}
While quantization make the inference of large models feasible on consumer GPUs, finding the optimal rotation still requires substantial computational power. We address this by avoiding end-to-end fine-tuning. Since each multi-head attention and FFN is affected by  $\mR_1$ , end-to-end approaches like SpinQuant cannot optimize the rotation layer by layer, and directly optimizing  $\mR_1$  via gradient descent requires loading the entire model, which is memory-intensive. Although SpinQuant reduces training costs by eliminating the need to store weight gradients and states, it still requires loading the full model into GPU memory. Our approach uses layer-wise inference, which eliminates the need to load all the network weight on the GPU at once. Then we store the activations for each layer. The we optimize the rotation with a Kurtosis loss. This significantly lowers GPU requirements—at most, a single NVIDIA H100 (or A100) is needed for LLaMA 70B.

\begin{table*}[t]
    \caption{Comparison of different quantization methods across various models. All the results are for 4 bit quantization for Weight/Activation/KV-cache. Weights are quantized using GPTQ.}
    \centering
    \resizebox{\textwidth}{!}{

\begin{tabular}{c|ccc|ccc|ccc}
\hline
\multirow{2}{*}{\textbf{Method}}      & \multicolumn{3}{c|}{\textbf{Llama-2-7b}}                                                      & \multicolumn{3}{c|}{\textbf{Llama-2-13b}}                                                      & \multicolumn{3}{c}{\textbf{Llama-3-8b}}                                                       \\ \cline{2-10} 
                                      & \textbf{Wiki ($\downarrow$)} & \textbf{0-shot ($\uparrow$)} & \textbf{MMLU ($\uparrow$)} & \textbf{Wiki ($\downarrow$)} & \textbf{0-shot ($\uparrow$)} & \textbf{MMLU ($\uparrow$)} & \textbf{Wiki ($\downarrow$)} & \textbf{0-shot ($\uparrow$)} & \textbf{MMLU ($\uparrow$)} \\ \hline
16-bit                                & 5.5                          & 64.1                              & 42.1                       & 4.9                          & 66.5                              & 52.7                       & 6.1                             &  67.2                                &     63.2                       \\
        \midrule
GPTQ                                  & 9600.0s                      & 38.9                              & 23.8                       & 3120.0                       & 33.8                              & 24.8                       & 166.3                        & 39.8                              & 23.3                            \\
QuaRot                                & 6.2                          & 60.6                              & 32.3                       & 5.4                          & 64.7                              & 46.83                       &  8.50                        & 60.1                        & 47.4                           \\
SpinQuant                             & 6.0                          & 61.0                              & 34.8                       & 5.2                          & 64.8                              & 47.8                       & 7.4                          & 63.8                              & 56.2                          \\
Kurtail                               & \textbf{5.9 }                         & \textbf{61.3 }                             & \textbf{32.9    }                   & \textbf{5.2}                          & \textbf{65.2}                              &\textbf{ 49.1}                       &  \textbf{7.2}                          & \textbf{64.6 }                        &\textbf{ 57.3   }                         \\ \hline
\multirow{2}{*}{\textbf{Method}}      & \multicolumn{3}{c|}{\textbf{Llama-3-70b}}                                                      & \multicolumn{3}{c|}{\textbf{Llama-3.2-1b}}                                                     & \multicolumn{3}{c}{\textbf{Llama-3.2-3b}}                                                     \\ \cline{2-10} 
                                      & \textbf{Wiki ($\downarrow$)} & \textbf{0-shot ($\uparrow$)} & \textbf{MMLU ($\uparrow$)} & \textbf{Wiki ($\downarrow$)} & \textbf{0-shot ($\uparrow$)} & \textbf{MMLU ($\uparrow$)} & \textbf{Wiki ($\downarrow$)} & \textbf{0-shot ($\uparrow$)} & \textbf{MMLU ($\uparrow$)} \\ \hline
16-bit                                & 6.1                          & 67.1                              & 63.1                       & 9.75                         & 54.9                              & 37.9                       & 7.8                          & 62.7                              & 54.8                       \\
        \midrule
GPTQ                                  & 166.3                        & 39.8                              & 23.3                       & 108.9                        & 38.0                              & 24.9                       & 178.3                        & 40.3                              & 24.8                       \\
QuaRot                                & 8.52                         & 60.1                              & 47.5                       & 17.4                         & 49.0                              & 23.8                       & 10.1                         & 56.1                              & 42.0                       \\
SpinQuant                             & 7.4                          & 63.8                              & 56.2                       & 13.6                         & 48.8                              & 25.6                       & 9.2                          & 57.9                              & 44.2                       \\
Kurtail                               & \textbf{7.2}                          & \textbf{64.6}                              & \textbf{57.34 }                     & \textbf{12.9}                         & \textbf{50.1}                              & \textbf{27.2}                       & \textbf{9.0}                  & \textbf{59.0}                     & \textbf{47.8}              \\ \hline
\end{tabular}

    }
    
    \label{tab: main4bit}
\end{table*}

\section{Setup}
\label{app: setup}

\paragraph{Setup.} We developed \method using the Hugging Face library \citep{HuggingFace} integrated with the PyTorch framework \citep{pytorch} and for evaluation we used EleutherAI evaluation framework \citep{eval-harness}. For learning the transformation, we used 512 calibration samples for all models, except Mixteral and LLAMA 70B  for which we use 256 calibration sample from the WikiText \citep{merity2016wikitext} training set, each with a sequence length of 2048. For large models, we used less samples since they have more layers for which we can store the activations. For storing the activations we used layer-wise inference to reduce the GPU memory requirement. For optimizing the rotation, we use Caley Adam \citep{li2020efficient} optimizer to find the rotation. 
For quantizing the activation, we used per-token dynamic symmetric quantization, where a single scale was applied to each row, and all values were clipped using a quantile of 0.98 in all experiments. For the KV-caches, we employed asymmetric quantization. For the Weight quantization, we use round-to-nearest (RTN), and GPTQ \citep{gptq}, using per-column (or per-channel) symmetric quantization. For GPTQ quantization, we uses 128 calibration samples from the WikiText, each with a sequence length of 2048. Learning the transformation and Transforming LLAMA3-70B with \method on an NVIDIA H100 GPU took around one hour which compare the SpinQuant it uses significantly less memory (4 A100 GPU and 2 hours). 

\paragraph{Models} We evaluate \method on the LLAMA-2 \citep{touvron2023llama}, LLAMA-3 \citep{grattafiori2024llama3herdmodels}, Phi-model family \citep{abdin2024phi3technicalreporthighly} on
both language generation and zero-shot tasks. We further also target the mixture of experts model Mixtral \citep{mixtral}.

\paragraph{Inference Speed-up.}
 KurTail's contribution focuses on a novel approach to learning the rotation and given the architectural similarity with SpinQuant and Quarot, we did not re-implement the low-level kernel for 4-bit matrix multiplication, as similar speedup results are expected. All results are based on simulated quantization; however, the real quantization will yield the same downstream performance.

\paragraph{Evaluation Setting.} To compare the performance of the model after quantization, we report the perplexity (PPL) score on the WikiText \citep{merity2016pointer} test set. While perplexity is a standard measure of language modeling performance, it may not be sufficient for evaluating the model’s effectiveness after quantization. Therefore, we report the result for zero-shot reasoning as well. We assess performance using the lm-evaluation-harness \citep{gao2024lessons}, testing the models on eight tasks: BoolQ \citep{clark2019boolq}, HellaSwag \citep{zellers2019hellaswag}, OpenBookQA(OBQA) \citep{mihaylov2018openbookqa}, PIQA \citep{bisk2020piqa}, SIQA \citep{sap2019socialiqa}, WinoGrande \citep{sakaguchi2021winogrande}, ARC-Easy, and ARC-Challenge \citep{boratko2018arc} reporting the average performance across all eight tasks (0-shot), we also provide the performance on each task in \autoref{app: further_eval}. Additionally, to assess the model on more complex tasks, we benchmark its language comprehension and general understanding using the MMLU benchmark \citep{hendrycks2021mmlu} and for mathematical reasoning we utilize MathQA \citep{amini2019mathqa}. We report the average performance in \autoref{tab: main4bit}.

\section{Results} To evaluate \method we focus on 4-bit quantization for weights, activations and KV-cache, which is a challenging bit-width for LLM quantization. \autoref{tab: main4bit} shows a summary where "0-shot" means the average performance over 8 tasks of common sense reasoning. For weight quantization we used GPTQ \citep{gptq}. We report the detailed performance of each tasks in \autoref{app: further_eval}. To demonstrate that our method outperforms previous works independently of the weight quantization technique, we alos provide results for round-to-nearest (RTN) in \autoref{app: further_eval}. Additionally, to show that our method is effective on LLM families beyond the LLaMA family, we present results on the Phi-3 model in \autoref{tab:phi3}.

\begin{table}[ht]
    \centering
    \caption{Performance on Phi-3-mini-4k-instruct.}
    \label{tab:phi3}
    \begin{tabular}{lccc}
        \toprule
         \textbf{Method} & \textbf{Wiki($\downarrow$)} & \textbf{0-shot($\uparrow$)} & \textbf{MMLU($\uparrow$)} \\
        \midrule
         16 bit  & 6.01 & 0.69 & 70.75 \\
         \midrule
         Quarot   & 8.46 & 0.61 & 56.01 \\
         \method  & \textbf{7.13} & \textbf{0.66} & \textbf{63.61} \\

        \bottomrule
    \end{tabular}
\end{table}

For all of the result we have better perplexity in all of the models compared to previous methods. At the same time, our method is significantly better that SpinQuant and QuaRot in downstream tasks. We provide further results for mixture of experts models in \autoref{app: mixexpert}. We also provide results for math reasoning in \autoref{app: math_reasoning}.

\subsection{Experiment on Mixture of Experts}
\label{app: mixexpert}
Given the growing popularity of the Mixture of Experts (MoE) models, we also explore the idea of applying rotation within the mixture of experts. For this purpose, we utilize Mixtral \citep{mixtral}, which employs the exact same attention block. However, for the mixture of experts component, we apply rotation across all the experts. \autoref{tab:mixexpert} presents the results for 4-bit quantization, where we used rounding to the nearest value. In principle, other quantization methods, such as GPTQ, HQQ \citep{badri2023hqq}, and similar approaches, can also be employed to further enhance performance.

\begin{table}[ht]
    \centering
    \caption{Performance comparison of different quantization methods for Mixtral-8x7B. All results correspond to 4-bit quantization for weights, activations, and KV-cache. RTN is used for weight quantization.}
    \label{tab:mixexpert}
    
    \begin{tabular}{l c c c}
        \toprule
        \multirow{2}{*}{\textbf{Method}} & \multicolumn{3}{c}{\textbf{Mixtral-8x7B}} \\ 
        \cmidrule(lr){2-4}
         & \textbf{Wiki ($\downarrow$)} & \textbf{0-shot ($\uparrow$)} & \textbf{MMLU ($\uparrow$)} \\ 
        \midrule
        16-bit     & 3.8  & 71.2  & 68.8 \\ 
        \midrule
        RTN        & 909.0 & 35.4  & 23.0 \\
        QuaRot     & 8.7   & 55.7  & 36.8 \\
        Kurtail    & \textbf{6.5}   & \textbf{59.4}  &\textbf{ 44.8} \\
        \bottomrule
    \end{tabular}
    
\end{table}

\begin{table}[ht]
    \centering
    \caption{Comparison of different quantization methods across various LLaMA model families and the Phi-3 model for mathematical reasoning on MathQA. All results are reported for 4-bit quantization of weights, activations, and KV cache. For weight quantization, we use GPTQ.}
    \label{tab:mathqa}
  \begin{tabular}{l| c c c}
    \toprule
        \multirow{2}{*}{\textbf{Model}} & \multicolumn{3}{c}{\textbf{MathQA Acc (\%)}} \\ 
    \cmidrule(lr){2-4}
      & \textbf{16-bit} & \textbf{QuaRot} & \textbf{\method} \\ 
    \midrule
    \midrule
    LLaMA-2-7B  & 28.24  & 26.70  & \textbf{26.77} \\
    LLaMA-2-13B & 31.76  & 28.81  & \textbf{30.35} \\
    LLaMA-2-70B & 38.39  & 33.97  & \textbf{35.68} \\
    \midrule
    \midrule
    LLaMA-3-8B  & 40.30  & 31.36  & \textbf{34.71} \\
    LLaMA-3-70B & 51.79  & 35.54  & \textbf{45.76} \\
    \midrule
    \midrule
    LLaMA-3.2-1B & 28.94  & 25.29  & \textbf{26.00} \\
    LLaMA-3.2-3B & 34.67  & 30.75  & \textbf{30.52} \\
    \midrule
    \midrule
    Phi-3-mini   & 39.93  & 31.89  & \textbf{34.81} \\
    \bottomrule
    \end{tabular}
\end{table}

\subsection{Evaluating Mathematical Reasoning}
\label{app: math_reasoning}
To explore more complex reasoning tasks, we further evaluate the performance of the quantized model on tasks involving mathematical reasoning in \autoref{tab:mathqa} by  reporting results on the MathQA \citep{amini2019mathqa} dataset. MathQA is a benchmark designed to test problem-solving and quantitative reasoning abilities. The dataset consists of real-world mathematical problems covering topics such as arithmetic, algebra, probability, and geometry. Each problem is accompanied by a natural language description, multiple-choice answers, and an annotated solution program that outlines the reasoning steps required to reach the correct answer. In \autoref{tab:mathqa}, we compare \method with QuaRot, and the results show that \method outperforms QuaRot. This additional observation suggests that optimizing the rotations can also enhance performance on math reasoning tasks.

\subsection{Ablation Study on the Calibration Dataset}
\label{sec: cal_abl}
We also investigate the impact of the calibration dataset on performance. To this end, we modify the calibration data to optimize the rotation using different datasets. Specifically, we conduct experiments using PTB \citep{marcus1993building}, C4 \citep{raffel2020exploring}, WikiText \citep{merity2016pointer}, and Alpaca \citep{taori2023stanford}. Additionally, we create a combined dataset by sampling equally from all four sources. For each experiment, we sample 512 instances and report the results for Llama-3.2 3B.

\begin{table}[ht]
    \centering
    \caption{Performance metrics on different calibration datasets.}
    \label{tab:calibration_results}
    \begin{tabular}{lccc}
        \toprule
        \textbf{Cal Dataset} & \textbf{Wiki($\downarrow$)} & \textbf{0-shot($\uparrow$)} & \textbf{MMLU($\uparrow$)} \\
        \midrule
        Quarot &  10.1 & 56.1  & 42.0 \\
        \midrule
        Wikitext-2  & \textbf{9.0} & 59.05 & 47.76 \\
        C4          & 9.1 & 59.24 & 47.75 \\
        Alpaca      & 9.3 & 59.68 & 47.34 \\
        PTB         & 9.2 & 58.60 & 48.33 \\
        Combined     & \textbf{9.0} & \textbf{59.79} & \textbf{48.75} \\

        \bottomrule
    \end{tabular}
\end{table}
\autoref{tab:calibration_results} presents the findings. Interestingly, all dataset variations outperform the non-training method Quarot. Moreover, we observe lower perplexity on WikiText when using other datasets for calibration. The best performance on the MMLU task is achieved with the PTB dataset, while the best results for common sense reasoning tasks are obtained using the Alpaca dataset. The combined dataset yields the best overall performance across all tasks while it uses the exact same number of samples (512 sentences).

In \autoref{tab:calsize_results}, we explore different calibration sample sizes for learning the rotations and their impact on the model’s performance in downstream tasks. In this study, we used our combined dataset and the Llama 3.2 3B model. As shown in \autoref{tab:calibration_results}, we observe a trend toward improvement as the sample size increases, although performance tends to saturate around a sample size of 512.

\begin{table}[ht]
    \centering
    \caption{Effect of different calibration size on quantization performance.}
    \label{tab:calsize_results}
    \begin{tabular}{cccc}
        \toprule
        \textbf{Cal Size} & \textbf{Wiki($\downarrow$)} & \textbf{0-shot($\uparrow$)} & \textbf{MMLU($\uparrow$)} \\
        \midrule
        128  & 9.11 & 59.24  & 47.85 \\
        256  & 9.12 & 58.85  & 47.47 \\
        512  & 9.09 & \textbf{59.79}  & 48.75 \\
        1024 & \textbf{9.08} & 59.43  & \textbf{49.02} \\
        \bottomrule
    \end{tabular}
\end{table}
\section{Conclusion}
We introduced \method -- a novel technique for learning orthogonal transformations that rotate the activation distribution to address the outlier problem. \method effectively reduces quantization sensitivity and minimizes quantization error by tackling important challenges, such as the outlier issue, and overcomes the limitations of previous approaches. Compared to QuaRot, which uses non-learnable rotation, and SpinQuant, which requires substantial computational resources for learning rotations, \method provides a more efficient and robust solution. 
These results highlight \method's ability to deliver efficiency and high performance across large-scale language models.

\section*{Limitations}

In this work, we only focuses on dynamic per-token quantization for activations, which offers flexibility but does not fully exploit the potential of static tensor-wise quantization. Static quantization, which precomputes scaling factors for improved efficiency, could further optimize inference speed and memory usage. However, it requires careful calibration, which we leave for future work.


\bibliography{kurtosis.bib}

\newpage
\appendix

\section{Further Evaluation}
\label{app: further_eval}
In this section, we provide a more detailed evaluation of all tasks. We present results for 4-bit quantization of weights, activations, and the KV-cache. \autoref{tab:mmlu_model_performance} reports the performance of each MMLU task under 4-bit quantization for weights, activations, and the KV-cache. We use the GPTQ quantization algorithm for weight quantization in this experiment. Similarly, using the same setup, we evaluate common-sense reasoning tasks, as shown in \autoref{tab:cr_gptq_model_performance}. Finally, we report the performance of common-sense reasoning tasks using RTN quantization for weights in \autoref{tab:cr_rtn_model_performance}.
\begin{table}[htbp]
    \centering
    \caption{Performance comparison of different models using various methods across different domains.}
    \label{tab:mmlu_model_performance}
    \resizebox{0.5\textwidth}{!}{
    \begin{tabular}{l||ccccc|c}
        \toprule
        \textbf{Model} & \textbf{Method} & \textbf{Human} & \textbf{Other} & \textbf{STEM} & \textbf{S-Sci} & \textbf{AVG} \\
        \midrule

        \multirow{4}{*}{Llama-2-7B}  
            & Vanilla   & 39.8 & 47.3 & 34.2 & 47.3 & 42.1 \\ 
            \cline{2-7}
            & Quarot    & 31.1 & 35.7 & 29.9 & 34.1 & 32.7 \\ 
            & SpinQuant & 33.9 & 38.5 & 29.5 & 37.5 & 34.8 \\
            & Kurtail   & 32.3 & 35.0 & 29.8 & 34.4 & 32.9 \\ 
        \midrule
        
        \multirow{4}{*}{Llama-2-13B}  
            & Vanilla   & 47.9 & 59.4 & 42.3 & 61.2 & 52.7 \\ 
            \cline{2-7}

            & Quarot    & 42.7 & 52.3 & 38.2 & 54.1 & 46.8 \\ 
            & SpinQuant & 43.5 & 53.1 & 39.1 & 55.4 & 47.8 \\
            & Kurtail   & 45.3 & 54.0 & 40.4 & 56.6 & 49.1 \\ 
        \midrule

        \multirow{4}{*}{Llama-3-8B}  
            & Vanilla   & 55.0 & 70.8 & 53.7 & 73.2 & 63.2 \\ 
            \cline{2-7}
            & Quarot    & 42.1 & 52.9 & 39.8 & 54.9 & 47.4 \\ 
            & SpinQuant & 49.8 & 63.3 & 46.8 & 65.0 & 56.2 \\
            & Kurtail   & 50.2 & 64.5 & 49.1 & 65.6 & 57.3 \\  
        \midrule
        
        \multirow{4}{*}{Llama-3-70B}  
            & Vanilla   & 67.7 & 81.5 & 69.2 & 86.7 & 76.3 \\
            \cline{2-7}
            & Quarot    & 55.3 & 68.5 & 53.7 & 74.1 & 62.9 \\ 
            & SpinQuant & 50.7 & 67.0 & 51.9 & 68.1 & 59.4 \\
            & Kurtail   & 65.2 & 79.1 & 63.9 & 84.2 & 73.1 \\  
        \midrule
        
        \multirow{4}{*}{Llama-3.2-1B}  
            & Vanilla   & 35.3 & 41.3 & 33.9 & 41.3 & 38.0 \\ 
            \cline{2-7}
            & Quarot    & 25.4 & 26.9 & 24.4 & 25.4 & 25.5 \\
            & SpinQuant & 25.4 & 27.6 & 24.2 & 25.3 & 25.6 \\
            & Kurtail   & 26.5 & 28.8 & 26.0 & 27.3 & 27.2 \\  
        \midrule
        
        \multirow{4}{*}{Llama-3.2-3B}  
            & Vanilla   & 49.0 & 63.1 & 45.5 & 62.9 & 55.1 \\ 
            \cline{2-7}
            & Quarot    & 38.5 & 47.3 & 35.3 & 46.7 & 42.0 \\
            & SpinQuant & 37.0 & 49.4 & 39.9 & 50.5 & 44.2 \\
            & Kurtail   & 44.8 & 53.4 & 39.5 & 53.4 & 47.8 \\  
        \bottomrule
    \end{tabular}
    }
\end{table}

\begin{table*}[ht]
    \centering
    \caption{Performance comparison of various models with 4 bits W/A/KV-cache quantization in common sense reasoning tasks. All the weight are quantized using GPTQ.}
    \label{tab:cr_gptq_model_performance}
    \resizebox{\textwidth}{!}{
    \begin{tabular}{l|c|cccccccc|c}
        \toprule
        \textbf{Model} & \textbf{Method} & \textbf{ARC-C} & \textbf{ARC-E} & \textbf{BoolQ} & \textbf{HellaSwag} & \textbf{OBQA} & \textbf{PIQA} & \textbf{SIQA} & \textbf{WinoGrande} & \textbf{AVG} \\
        \midrule \midrule
        
        \multirow{4}{*}{Llama-2-7B}  
            & Vanilla   & 46.2 & 74.5 & 77.8 & 76.0 & 44.2 & 79.1 & 46.1 & 69.1 & 64.1 \\
            \cdashline{2-11} 
            & Quarot    & 41.6 & 70.6 & 73.2 & 72.1 & 41.2 & 76.9 & 44.0 & 65.2 & 60.6 \\
            & SpinQuant & 43.6 & 71.3 & 73.8 & 73.2 & 40.4 & 76.0 & 44.1 & 65.4 & 61.0 \\
            & Kurtail   & 43.1 & 72.0 & 72.0 & 73.2 & 41.2 & 76.6 & 45.6 & 66.8 & 61.3 \\
        \midrule
        \multirow{4}{*}{Llama-2-13B}  
            & Vanilla   & 49.2 & 77.5 & 80.6 & 79.4 & 45.2 & 80.5 & 47.4 & 72.1 & 66.5 \\
            \cdashline{2-11} 

            & Quarot    & 47.3 & 73.9 & 77.8 & 76.6 & 44.4 & 78.7 & 44.1 & 69.8 & 64.1 \\
            & SpinQuant & 49.0 & 76.3 & 78.2 & 77.1 & 42.8 & 79.3 & 46.3 & 69.5 & 64.8 \\
            & Kurtail   & 48.1 & 75.4 & 79.7 & 77.4 & 45.0 & 79.0 & 45.6 & 71.2 & 65.2 \\
        \midrule
        \multirow{4}{*}{Llama-3-8B}  
            & Vanilla   & 53.4 & 77.8 & 81.4 & 79.2 & 45.0 & 80.8 & 47.2 & 72.6 & 67.2 \\
            \cdashline{2-11} 

            & Quarot    & 42.1 & 69.0 & 72.1 & 71.5 & 41.2 & 74.9 & 44.3 & 65.5 & 60.1 \\
            & SpinQuant & 48.0 & 75.4 & 75.8 & 75.4 & 43.8 & 77.5 & 45.0 & 69.2 & 63.8 \\
            & Kurtail   & 48.2 & 75.4 & 79.2 & 76.4 & 43.6 & 78.4 & 45.8 & 70.0 & 64.6 \\
        \midrule
        \multirow{4}{*}{Llama-3-70B}  
            & Vanilla   & 65.0 & 86.6 & 85.4 & 85.0 & 48.2 & 84.3 & 50.5 & 79.9 & 73.1 \\
            \cdashline{2-11} 
            & Quarot    & 53.0 & 74.8 & 81.2 & 77.7 & 42.0 & 78.2 & 45.7 & 68.4 & 65.1 \\
            & SpinQuant & 52.0 & 77.3 & 81.7 & 75.6 & 43.8 & 78.8 & 43.4 & 72.8 & 65.7 \\
            & Kurtail   & 59.2 & 82.7 & 83.9 & 83.3 & 46.6 & 83.5 & 49.7 & 76.6 & 70.7 \\
        \midrule
        \multirow{4}{*}{Llama-3.2-1B}  
            & Vanilla   & 36.2 & 60.4 & 63.9 & 63.6 & 37.2 & 74.6 & 43.0 & 60.5 & 54.9 \\
          \cdashline{2-11} 

            & Quarot    & 30.0 & 51.4 & 59.1 & 54.0 & 34.2 & 66.7 & 39.6 & 57.1 & 49.0 \\
            & SpinQuant & 32.3 & 51.8 & 59.3 & 55.4 & 30.4 & 67.7 & 38.6 & 54.7 & 48.8 \\
            & Kurtail   & 31.1 & 52.9 & 60.7 & 56.4 & 36.4 & 68.6 & 40.5 & 54.3 & 50.1 \\
        \midrule
        \multirow{4}{*}{Llama-3.2-3B}  
            & Vanilla   & 46.0 & 71.7 & 73.2 & 73.6 & 43.0 & 77.5 & 47.0 & 69.7 & 62.7 \\
            \cdashline{2-11} 

            & Quarot    & 38.6 & 59.0 & 65.9 & 66.5 & 35.8 & 74.4 & 43.1 & 65.2 & 56.1 \\
            & SpinQuant & 38.9 & 64.8 & 68.0 & 69.1 & 39.4 & 74.9 & 45.1 & 62.9 & 57.9 \\
            & Kurtail   & 42.2 & 66.7 & 69.8 & 68.8 & 39.8 & 75.6 & 44.8 & 64.6 & 59.0 \\
        \bottomrule
    \end{tabular}

    }

\end{table*}

\begin{table*}[ht]
\centering
\caption{Performance comparison of various models with 4 bits W/A/KV-cache quantization in common sense reasoning tasks. All the weights are quantized using RTN.}
\label{tab:cr_rtn_model_performance}
\resizebox{\textwidth}{!}{
\begin{tabular}{l|c|cccccccc|c}
        \toprule
        \textbf{Model} & \textbf{Method} & \textbf{ARC-C} & \textbf{ARC-E} & \textbf{BoolQ} & \textbf{HellaSwag} & \textbf{OBQA} & \textbf{PIQA} & \textbf{SIQA} & \textbf{Winogrande} & \textbf{AVG} \\
        \midrule \midrule
        \multirow{3}{*}{Llama-2-7B}   & Vanilla  & 46.2 & 74.5 & 77.8 & 76.0 & 44.2 & 79.1 & 46.1 & 69.1 & 64.1 \\
                \cdashline{2-11} 

                                      & Quarot   & 35.2 & 62.4 & 69.0 & 62.6 & 33.4 & 71.7 & 40.9 & 60.2 & 54.4 \\
                                      & Kurtail  & 39.0 & 64.9 & 69.8 & 64.7 & 39.2 & 74.1 & 42.1 & 62.2 & 57.0 \\
        \midrule
        \multirow{3}{*}{Llama-2-13B}  & Vanilla  & 49.2 & 77.5 & 80.6 & 79.4 & 45.2 & 80.5 & 47.4 & 72.1 & 66.5 \\
                    \cdashline{2-11} 

                                      & Quarot   & 41.4 & 68.2 & 73.2 & 71.2 & 41.6 & 76.3 & 41.1 & 66.1 & 59.9 \\
                                      & Kurtail  & 44.2 & 70.3 & 74.7 & 72.5 & 40.4 & 77.5 & 45.9 & 70.2 & 62.0 \\
        \midrule
        \multirow{3}{*}{Llama-2-70B}  & Vanilla  & 57.4 & 81.1 & 83.8 & 83.8 & 48.8 & 82.8 & 49.2 & 78.0 & 70.6 \\
                    \cdashline{2-11} 

                                      & Quarot   & 50.5 & 76.8 & 80.0 & 78.4 & 44.0 & 79.9 & 46.0 & 72.9 & 66.1 \\
                                      & Kurtail  & 51.3 & 76.6 & 80.9 & 81.0 & 46.4 & 81.7 & 46.8 & 76.2 & 67.6 \\
        \midrule
        \multirow{3}{*}{Llama-3-8B}   & Vanilla  & 53.4 & 77.8 & 81.4 & 79.2 & 45.0 & 80.8 & 47.2 & 72.6 & 67.2 \\
                    \cdashline{2-11} 

                                      & Quarot   & 31.1 & 51.6 & 55.7 & 62.0 & 31.6 & 66.3 & 40.1 & 59.0 & 49.7 \\
                                      & Kurtail  & 38.1 & 61.1 & 72.5 & 69.3 & 36.8 & 72.9 & 41.9 & 66.1 & 57.3 \\
        \midrule
        \multirow{3}{*}{Llama-3-70B}  & Vanilla  & 65.0 & 86.6 & 85.4 & 85.0 & 48.2 & 84.3 & 50.5 & 80.0 & 73.1 \\
                    \cdashline{2-11} 

                                      & Quarot   & 20.6 & 31.3 & 58.5 & 28.4 & 25.4 & 55.0 & 33.2 & 50.7 & 37.9 \\
                                      & Kurtail  & 23.0 & 37.8 & 48.5 & 33.9 & 29.8 & 61.8 & 36.6 & 51.6 & 40.4 \\
        \midrule
        \multirow{3}{*}{Llama-3.2-1B} & Vanilla  & 36.2 & 60.4 & 63.9 & 63.6 & 37.2 & 74.6 & 43.0 & 60.5 & 54.9 \\
                    \cdashline{2-11} 

                                      & Quarot   & 27.4 & 33.9 & 39.1 & 36.2 & 30.0 & 56.9 & 34.7 & 53.0 & 38.9 \\
                                      & Kurtail  & 28.7 & 37.2 & 38.8 & 42.9 & 31.6 & 60.0 & 35.7 & 57.5 & 41.5 \\
        \midrule
        \multirow{3}{*}{Llama-3.2-3B} & Vanilla  & 46.0 & 71.7 & 73.2 & 73.6 & 43.0 & 77.5 & 47.0 & 69.7 & 62.7 \\
                    \cdashline{2-11} 

                                      & Quarot   & 33.1 & 50.3 & 41.8 & 56.3 & 31.8 & 67.8 & 39.8 & 56.8 & 47.2 \\
                                      & Kurtail  & 37.4 & 56.6 & 48.0 & 62.1 & 36.6 & 71.3 & 40.5 & 60.4 & 51.6 \\

        \bottomrule
    \end{tabular}
}

\end{table*}

\end{document}